%% file: main.tex
\pdfoutput=1
\documentclass[11pt]{article}

\usepackage[]{EMNLP2023}

\usepackage{times}
\usepackage{latexsym}
\usepackage{booktabs}
\usepackage[T1]{fontenc}
\usepackage[inline]{enumitem}
\usepackage[utf8]{inputenc}

\usepackage{microtype}
\usepackage{multirow}
\usepackage{makecell}
\usepackage{diagbox}

\usepackage{graphicx}
\usepackage{tikz}
\usepackage{standalone}

\usepackage{pgfplots}
\pgfplotsset{compat=1.16}
\usepgfplotslibrary{statistics}
\usepgflibrary{plotmarks}
\usetikzlibrary{calc, patterns}

\usepackage{inconsolata}
\usepackage{amssymb}
\usepackage{amsmath}
\usepackage{pifont}
\usepackage{xspace}

\usepackage{listings}

\usepackage[framemethod=TikZ]{mdframed}
\newcounter{theo}[section] 
\newenvironment{theo}[2][]{%
\refstepcounter{theo}%
\ifstrempty{#1}%
{\mdfsetup{%
frametitle={%
\tikz[baseline=(current bounding box.east),outer sep=0pt]
\node[anchor=east,rectangle,fill=darkblue!20]
{\strut Prompt~\thetheo};}}
}%
{\mdfsetup{%
frametitle={%
\tikz[baseline=(current bounding box.east),outer sep=0pt]
\node[anchor=east,rectangle,fill=darkblue!20]
{\strut Prompt~\thetheo:~#1};}}%
}%
\mdfsetup{innertopmargin=10pt,linecolor=darkblue!20,%
linewidth=2pt,topline=true,font=\small,%
frametitleaboveskip=\dimexpr-\ht\strutbox\relax,
backgroundcolor=lightgray!20
}
\begin{mdframed}[nobreak=true]\relax%
\label{#2}}{\end{mdframed}}

\newcommand{\cmark}{\ding{51}}%
\newcommand{\xmark}{\ding{55}}%

\newcommand{\datapve}{\textsc{PVE}\xspace}
\newcommand{\datapersp}{\textsc{Perspectrum}\xspace}
\newcommand{\datakp}{\textsc{ArgKP}\xspace}

\input{macros}

\title{An Empirical Analysis of Diversity in Argument Summarization}

\author{Michiel van der Meer\\
  LIACS \\ Leiden University \\
  \texttt{\small m.t.van.der.meer@liacs.leidenuniv.nl} \\\And
  Piek Vossen \\
  CLTL \\ Vrije Universiteit Amsterdam \\
  \texttt{\small p.t.j.m.vossen@vu.nl} \\\AND
  Catholijn M. Jonker \\
  Interactive Intelligence \\ TU Delft \\
  \texttt{\small c.m.jonker@tudelft.nl} \\\And
  Pradeep K. Murukannaiah \\    
  Interactive Intelligence \\ TU Delft \\
  \texttt{\small p.k.murukannaiah@tudelft.nl}}

\begin{document}
\maketitle
\begin{abstract}
Presenting high-level arguments is a crucial task for fostering participation in online societal discussions. Current argument summarization approaches miss an important facet of this task---capturing \emph{diversity}---which is important for accommodating multiple perspectives. We introduce three aspects of diversity: those of opinions, annotators, and sources. 
We evaluate approaches to a popular argument summarization task called Key Point Analysis, which shows how these approaches struggle to (1) represent arguments shared by few people, (2) deal with data from various sources, and (3) align with subjectivity in human-provided annotations. We find that both general-purpose LLMs and dedicated KPA models exhibit this behavior, but have complementary strengths. Further, we observe that diversification of training data may ameliorate generalization. Addressing diversity in argument summarization requires a mix of strategies to deal with subjectivity.
\end{abstract}

\section{Introduction}
Getting an overview of the arguments concerning controversial issues is often difficult for those participating in ongoing discussions. This is because there are many points being communicated, no way to track which arguments were already encountered, and haphazard miscommunication or conflict.
Automatic summarization is a way to provide a comprehensible overview of the opinions \citep{nenkova2011automatic, angelidis2018summarizing}. However, generating summaries representative of the arguments involved in a discussion is difficult \citep{bar2020arguments}. Argument summarization extends beyond text summarization because it separates argumentative and non-argumentative content, preserves the argumentative structure, and provides explicit stances on a central claim or hypothesis. 

Summarizing arguments is challenging in many contexts, but the potential impact is high. For instance, after summarizing the arguments from societal discussions, the extracted arguments may shape new policies and be used to justify decision-making \citep{arana2021citizen, gurkan2010mediating}. Similarly, businesses depend on review data to find customer feedback, which can be used to steer product design \citep{archak2007show}. 

Although arguments are often summarized by hand in practice \citep[e.g.,][]{mouter2021public, mclaren2016public, nahm2013novel}, recent developments in Argument Mining (AM) allow automatic analysis of argumentative text \citep{lawrence2020argument}. Obtaining summaries that faithfully represent open-ended opinions requires careful evaluation, especially in sensitive contexts, e.g., summarizing citizen feedback \citep{egan2016summarising, misra2015using}. 

One approach for generating comprehensive summaries of arguments is Key Point Analysis \citep[KPA,][]{bar2020arguments}. In KPA, a corpus of opinions is analyzed for the \emph{key points}, those arguments that are salient and repeated multiple times. However, some aspects of the KPA experimental design misalign with respect to real-world applications. 
We illustrate these blind spots, in particular, when applied to summarizing online societal discussions. We highlight three dimensions of \textbf{diversity} that are central to empowering citizens' opinions at scale \citep{shortall2022reason}:
\begin{enumerate*}[label=(\arabic*)]
    \item incorporating the long tail of opinions,
    \item  being robust in handling data from multiple sources, and 
    \item including diverse perspectives from annotators.
\end{enumerate*}

How current KPA approaches deal with the above dimensions of diversity is unexplored. We conduct an empirical study of different argument summarization approaches by incorporating the standardized benchmark and two other datasets to experiment. 
We develop specific analyses to uncover how KPA approaches fare on each dimension of diversity. In addition to the existing approaches, we use LLMs by prompting them to perform KPA, as they may be an attractive alternative to current models. 

Applying KPA approaches across several datasets that vary in how they address diversity leads to mixed results. KPA performance degrades when dealing with low-frequency opinions, i.e., opinions repeated by relatively few individuals. Further, we observe that KPA approaches disregard subjective interpretations among individual annotators. Finally, they generalize poorly across data sources when used in transfer learning settings, though approaches reveal complementary merits across tasks.

\paragraph{Contributions}
\begin{enumerate*}[label=(\arabic*)]
    \item We critically examine three dimensions of diversity---of opinions, sources, and annotators---in the KPA setup. 
    \item We analyze the behavior of existing metrics on one existing and two newly adapted datasets. 
    \item We analyze multiple methods, including prompt-based LLMs, broadening the scope of methods that can perform KPA. 
\end{enumerate*}

\section{Related Work}
We outline three lines of related work: key point analysis, opinion summarization, and diversity. % in societal decision-making. 

\subsection{Key Point Analysis}
KPA serves to separate argumentative from non-argumentative content, and condense argumentative content by matching arguments to key points \citep{bar2020arguments}. Key points can be seen as high-level arguments that capture the gist of a set of arguments. While most work on KPA selects high-quality arguments as representatives, generating novel key points has been proposed as an alternative \citep{syed2021generating}. KPA has been applied across topics using data from discussion portals or online reviews \citep{bar2020quantitative, bar2021every}. KPA is usually divided into Key Point Generation and Key Point Matching steps (see Section~\ref{sec:kpa-task}). 

Multiple approaches exist for KPA \citep{friedman2021overview}. Modeling choices consist of popular Transformer models such as BERT \citep{phan2021matching}, enhanced representational quality using contrastive learning \citep{alshomary2021key}, and the incorporation of clustering techniques \citep{li2023hear}. Our work aims to investigate some of the modeling choices employed in these works. For instance, in \citet{li2023hear}, the authors discarded unmapped arguments, which may hurt the ability the represent minority opinions. 

\subsection{Opinion Summarization}
Opinion summarization aims to generate summaries of an individual's subjective opinions \citep{inouye2011comparing, bhatia2020comparative}, often applied to product reviews \citep{chu2019meansum}. Leveraging Transformer models is popular for opinion summarization \citep{angelidis2021extractive, amplayo2021aspect}, though generic extractive summarization techniques are strong baselines \citep{suhara2020opiniondigest}. Measuring bias in generated summaries has seen recent interest, specifically acknowledging that diverse opinions should be taken into account \citep{huang2023examining, siledar2023aspect} \arrup{or postulating that diversity is a desirable trait when generating opinions \citep{alshomary2022generating, wang2016neural}}. \arrup{Our work applies these techniques to argumentation to obtain a high-level summary of opinions, and analyses differences in behavior for (in-)frequent viewpoints. }

\subsection{Diversity in Societal Decision Making}
Sensitive decision-making contexts call for responses rooted in reason that serve social good rather than specific interests. One way of obtaining such responses is through evidence-based policymaking, which involves stakeholders and the broader public to strike decisions \citep{cairney2016politics}. Citizen participation improves the support of the decisions when some requirements are met \citep{mansbridge2012systemic}. A key factor among those requirements is the involvement of a diverse group of citizens, independently voicing opinions \citep{surowiecki2005wisdom}. Approaches to summarizing arguments in such citizen feedback face similar requirements. 

%Concerning the field of Argument Mining, we find few efforts that explicitly align with these views. 
In Argument Mining, we find recent work that aligns with these views, e.g., by a strong focus on the diverging perspectives among annotators in AM tasks \citep{romberg2022your}. Further, some preliminary work adjusts visualization for minority opinions \citep{baumer2022course}. However, in terms of data sources, most work is still centered on English-speaking content, with few multi-lingual or multi-cultural resources available \citep{vecchi2021towards}.
%such as those under strict time constraints or with

\section{Method}
We formulate the KPA subtasks---\emph{Key Point Generation} (KPG) and \emph{Key Point Matching} (KPM). %---and observe their performance separately to uncover the strengths and weaknesses of each method on each task. 
We then introduce the three dimensions of diversity and consider them when applying KPA. % to existing resources. 

\begin{table*}[htb]
    \centering
    \begin{tabular}{@{}llcccc@{}}
        \toprule
        \textbf{Dataset} & \textbf{Data Source} & \makecell[c]{\textbf{Filter}\\\textbf{low freq.}} & \textbf{\makecell[c]{\textbf{Key Points}\\ \textbf{source}}} & \makecell[c]{\textbf{Non-aggregated}\\\textbf{annotation}} & \textbf{IRR} \\
        \midrule
        \datakp & Human annotation & \cmark  & Expert & \xmark & 0.50-0.82 ($\kappa$) \\
        \datapve & Citizen consultation & \xmark & Crowd & \cmark & 0.35 ($\kappa^{\dagger}$) \\
        \datapersp & Debate platforms & \xmark & Crowd & \xmark & 0.61 ($\kappa$)\\
        \bottomrule
    \end{tabular}
    \caption{Datasets and their diversity characteristics when considering the KPA task. The inter-rater reliability (IRR) is measured via Cohen's $\kappa$ scores or prevalence and bias-adjusted Cohen's $\kappa^{\dagger}$ \citep[PABAK,][]{sim2005kappa}. }
    \label{tab:datasets-alignment}
\end{table*}

\subsection{Task setup}
\label{sec:kpa-task}
We outline the two subtasks that constitute KPA, as originally introduced by \citep{friedman2021overview}. % which takes as input a corpus of comments and outputs a set of key points, matching comments to key points.

\begin{description}[itemsep=0pt, leftmargin=0pt]
\item[Key Point Generation (KPG)] focuses on generating \emph{key points} $\mathcal{K}$ given a corpus of arguments $\mathcal{D}$ on a particular claim. Key points are high-level arguments that capture the gist of a collection of arguments. Key points oppose or support the claim.

\item[Key Point Matching (KPM)] \emph{matches} arguments to key points. An argument matches a key point if the key point directly summarizes the argument, or if the key point represents the essence of the argument. We ensure that the stance of the key point (pro or con) matches the stance of the argument. Formally, given a set of key points $\mathcal{K}$ and a corpus $\mathcal{D}$, we score the match between an argument $d \in \mathcal{D}$ and a key point $k \in \mathcal{K}$ with a matching model $M(d,k)$. Assigning arguments to key points using match scores is flexible, and multiple strategies can be taken to reach a final decision (e.g. imposing a match score threshold) \citep{bar2020quantitative}. Since the assignment strategy is largely context-dependent, we evaluate the scoring mechanism itself, instead. % of the individual assignment strategies.
\end{description}

\subsection{Modeling Diversity in Key Point Analysis}
\label{sec:diversity}
We focus on three main aspects of diversity. %pertaining to the KPA task, the models and data used, as well as the context in which they are applied. 

\paragraph{(1) Long tail opinions} Several NLP models imitate biases that exist in datasets \citep{blodgett2020language}. For argument summarization, a form of bias is focusing on majority arguments, leading to possible misrepresentations. Failing to capture low-frequency arguments runs the danger of further estranging underrepresented viewpoints \citep{klein2012enabling}. These methods need active correction from humans to account for this ``long tail of opinions'' \citep{van2022hyena}. For the KPA task, approaches have largely unknown behavior on capturing the long tail of opinions \citep{mustafaraj2011vocal}. Additionally, LLMs struggle with learning long-tail knowledge \citep{kandpal2023large}, aggravating this issue. We experiment with subsampling the datasets to investigate the imbalanced data settings, which are representative of real-world use cases. 

\paragraph{(2) Annotators} Datasets are labeled using a mix of crowd and expert annotators. Querying experts for key points may leave the impacted users (e.g., lay citizens) out of consideration \citep{basile2021toward}. Similarly, labels stemming from crowd annotation that are filtered for high agreement may disregard controversial or diverse opinions. Disagreement is a complex signal that includes subjective views, task understanding, and annotator behavior \citep{aroyo2015truth}. Having access to non-aggregated annotations would, e.g., allow for further modeling of patterns \citep{davani2022dealing} or the reasons \citep{liscio2023value} underlying opinions. We investigate whether models trained on such annotations can identify disagreement. 

\paragraph{(3) Data sources} Existing works investigate cross-domain generalization of KPA methods using data stemming from 1a single dataset, focusing on a cross-topic setting \citep{bar2020quantitative, samin2022arguments, li2023hear}. This dataset is gathered at a specific time. As discussions evolve, more nuanced positions may become relevant, and new real-world events impact the opinions. Further, these discussions usually take place on a single platform (e.g., Reddit threads, Twitter discussions), inheriting biases from the source \citep{hovy2021five}. Measuring the performance of KPA approaches should rely on diverse datasets, based on data gathered from different sources at different points in time. There have been some efforts in applying KPA across different contexts \citep{gretz2023benchmark, bar2021every, cattan2023key}, but they apply approaches to a single dataset at a time, making direct comparison difficult. %This makes direct comparison of strengths and weaknesses difficult.
\arrup{Our work examines the cross-dataset performance of these approaches to assess their relative strengths and weaknesses.}

Table~\ref{tab:datasets-alignment} shows the current datasets, and how they relate to the dimensions discussed above. In all three datasets, the arguments stem from user-submitted content. In one dataset, low-frequency arguments (i.e., opinions repeated by few individuals) are disregarded. Further, the \datakp benchmark relies on expert-generated key points and does not include annotator-specific match labels. \datapersp contains aggregated counts of match labels, but due to aggregation, we cannot identify annotator-specific patterns. Lastly, the inter-rater reliability differs for each dataset, with wide ranges, showing that the tasks are fundamentally subjective. We employ these three datasets for evaluating various KPA approaches and dive deeper into the three aspects of diversity.

\section{Experimental Setup}
We describe the data, KPA methods, and metrics involved in our experiments. 
We make our source code\footnote{\url{https://github.com/m0re4u/argument-summarization}} and data \citep{vandermeer2024empirical_data} publicly available.

\subsection{Data}
Most work on KPA has used the dataset introduced by \citet{friedman2021overview} in a shared task. We add two new datasets that match the KPA subtasks but have different characteristics. 

\begin{description}[itemsep=0pt, leftmargin=0pt]
    \item[ArgKP] We adopt the shared task dataset, keeping the same split across claims as the original data. The \datakp dataset contains claims taken from an online debate platform, together with crowd-generated arguments and expert-generated key points \citep{bar2020arguments}. The arguments were produced by asking humans to argue for and against a claim, followed by filtering on high-quality and clear-polarity arguments. Key points were generated by an expert debater, who generated the key points without having access to the arguments. The final test set was collected after the initial dataset and has been curated to match some of the distributional properties of the training and validation sets. 
    \item[\datapve] We use the crowd-annotated data stemming from a human-AI hybrid key argument analysis \citep{van2022hyena} based on a Participatory Value Evaluation (PVE), a type of citizen consultation. In this consultation process, citizens were asked to motivate their choices for new COVID-19 policy through text, which formed a set of comments for each proposed policy option. The performed key argument analysis resulted in crowd-generated key points, matching individual comments to key points per option. Since this is a small dataset, we only use it for evaluation.
    \item[Perspectrum] Similar to \datakp, \datapersp contains content from online debate platforms. It extracts claims, key points, and arguments from the platform directly \citep{chen2019seeing}. Part of the dataset is further enhanced by crowdsourcing paraphrased arguments and key points. The \datapersp dataset is ordered into claims, which are argued for or against by perspectives, with evidence statements backing up each perspective. We use perspectives as key points, and evidence as arguments. We retain the same split over claims as the original data. The authors provide aggregated annotations on the match between arguments and key points. While this allows us to compute the agreement scores per sample, we cannot distil individual annotator patterns. 
\end{description} 

\begin{table}[tb]
    \centering
    \begin{tabular}{@{}lccc@{}}
        \toprule
        \textbf{Dataset} & \textbf{Train} & \textbf{Val} & \textbf{Test} \\  
        \midrule
        \datakp & 24 (21K) & 4 (3K) & 3 (3K) \\
        \textsc{\datapve} & -- & -- & 3 (200) \\
        \textsc{Perspec.} & 525 (6K) & 136 (2K) & 218 (2K)\\
        \bottomrule
    \end{tabular}
    \caption{Number of claims (and arguments) when splitting the dataset into training, validation, and test sets. }
    \label{tab:datasets-splits}
\end{table}

\subsection{Approaches}
We investigate different approaches with respect to their performance on the aspects of diversity. Appendix~\ref{app:experimental-setup} includes a detailed overview of the setup, parameters, and prompts.
Similar to summarization techniques, most KPG methods are either \emph{extractive}, taking samples as representative key points, or \emph{abstractive}, formulating new key points as free-form text generation \citep{el2021automatic}. 
\begin{description}[itemsep=0pt, leftmargin=0pt]
    \item[ChatGPT] We use the OpenAI Python API \citep{openai-python-api-2023} to run the KPA task by prompting ChatGPT. We differentiate between open-book and closed-book prompts. For the open-book prompts, we input the claim and a random sequence of arguments up to the maximum window (given a response size of 512 tokens) in the KPG task. For the closed-book model, we only input the claim, and the model synthesizes key points. In both approaches, KPG is abstractive. In KPM, ChatGPT predicts matches for a batch of arguments at a time, all related to the same claim. 
    \item[Debater] We use the Project Debater API \citep{debater-api-2023}, which supports multiple argument-related tasks, including KPA \citep{barhaim2021project}. This approach uses a model trained on \datakp and performs extractive KPG. We query the API for KPG and KPM separately. 
    \item[SMatchToPR] We adopt the approach from the winner of the shared task, which uses a state-of-the-art Transformer model and contrastive learning \citep{alshomary2021key}. During training, the model learns to embed matching arguments closer than non-matching arguments. These representations are used to construct a graph with embeddings of individual argument sentences as nodes, and the matching scores between them as edge weights. Nodes with the maximum PageRank score are selected as key points. In our experiments, the model is trained using the training set of \datakp and \datapersp. This method performs extractive KPG. We experiment with RoBERTa-base and RoBERTa-large to estimate the effect of model size \citep{liu2019roberta}. 
\end{description}

\subsection{Evaluation Metrics}
We evaluate models for KPG and KPM separately. For KPG, we adopt the set-level evaluation approach from \citet{li2023hear}. For KPM, we reuse the match labels provided by each dataset. 

\subsubsection{KPG} KPG can be considered as a language generation problem \citep{gatt2018survey} for evaluation. We rely on a mixture of reference-based and learned metrics, measuring both lexical overlap and semantic similarity. We use the following metrics:
\begin{description}[nosep, leftmargin=0pt]
    \item [ROUGE-(1/2/L)] to measure overlap of unigrams, bigrams, and longest common subsequence, respectively. %We compare the generated key points to all references of the same stance and claim, taking the maximum scoring combination as the final score. 
    We average scores for all stance and claim combinations. Additional details on the ROUGE configuration are in Appendix~\ref{app:sec-rouge}.
    \item [BLEURT] \citep{sellam2020bleurt} to measure the semantic similarity between a candidate and reference key point, which correlates with human preference scores. BLEURT introduces a regression layer over contextualized representations, trained on a set of human-generated labels. 
    % We pick the soft-F1 score to balance between precision and recall. %-oriented evaluation of the semantic similarity.
    \item [BARTScore] \citep{yuan2021bartscore} to evaluate the summarization capabilities directly by examining key point generation. In contrast to BLEURT, BARTScore evaluates the likelihood of the generated sequence when conditioning on a source. % Again, we report the soft-F1 scores.  
\end{description}

For each metric $\mathcal{S}$ that scores the overlap between two key points, we aggregate scores into Precision $P$ and Recall $R$ scores using Equations~\ref{eq:rouge-p} and~\ref{eq:rouge-r}. 
For $P$, we take the maximum score between a generated key point $a$ and the reference key points $\mathcal{B}$, averaging over all $n = |\mathcal{A}|$ generated key points. We perform the analogous for $R$. 
We report $F_1$ scores to balance precision and recall. 

\begin{align}
    P &= \frac{1}{n} \sum_{a \in \mathcal{A}} \underset{b \in \mathcal{B}}{\max} \;\mathcal{S}(a, b)     \label{eq:rouge-p}\\
    R &= \frac{1}{m} \sum_{b \in \mathcal{B}} \underset{a \in \mathcal{A}}{\max} \;\mathcal{S}(a, b)     \label{eq:rouge-r}
\end{align}

\subsubsection{KPM} We perform the KPM evaluation by obtaining match scores for key point-argument pairs. That is, for a key point $k$ and an argument $d$, we check if a new model used in the KPA method would assign $d$ to $k$. We reuse existing labels and do not use the results from KPG. 
Since we do not consider unlabeled examples between arguments and key points, we do not need to distinguish for undecided labels (as in \citet{friedman2021overview}). 

We evaluate each approach using mean average precision (mAP), taking the mean over average precision scores computed for claims $C$. Given a claim, we compute precision $P_{\tau}$ and recall $R_{\tau}$ for all match score thresholds $\tau$, as in Equation~\ref{eq:map}. In case an approach outputs a binary match label instead of scores, we remap the scores to 0 and 1 for non-matching and matching pairs, respectively. 
\begin{equation}
\label{eq:map}
    \textnormal{mAP} = \sum_{C}\frac{\sum_{\tau} (R_{\tau} - R_{\tau-1})P_{\tau}}{|C|}
\end{equation}

\section{Results and Discussion}
First, we report on the KPG and KPM evaluation. Then, we analyze how the aspects of diversity impact performance beyond a cross-dataset evaluation. We show results when conditioning on the long tail of opinions, look into the connection between annotator agreement and match score, and how performance changes for diverse data sources. 

\subsection{KPG Performance}
\label{sec:results-kpg}
\begin{table*}[htb]
    \centering
    \begin{tabular}{@{}llccccc@{}}
         \toprule
        \textbf{Dataset} & \textbf{Approach} & \textbf{R-1} & \textbf{R-2} & \textbf{R-L} & \textbf{BLEURT} & \textbf{BART} \\
         \midrule
         \multirow{4}{*}{\datakp} & ChatGPT             & \textbf{34.3} & \textbf{12.5} & \textbf{30.3} & 0.556 & \textbf{0.540} \\
         & ChatGPT (closed book)                        & 29.5 & 7.1 & 25.6 & 0.314 & 0.256\\
         & Debater                                      & 25.6 & 5.5 & 22.5 & 0.334 & 0.307 \\
         & SMatchToPR (base)                            & 31.7 & 11.1 & 29.7 & 0.553 & 0.494\\
         & SMatchToPR (large)                           & 30.5 & 8.3 & 26.8 & \textbf{0.563} & 0.497\\
         \midrule
         \multirow{4}{*}{\datapve} &          ChatGPT   & 18.5 & 3.9 & 15.3 & 0.329 & 0.369\\
         & ChatGPT (closed book)                        & \textbf{27.1} & \textbf{8.6} & \textbf{21.4} & \textbf{0.376} & \textbf{0.378}\\
         & Debater                                      & 13.3 & 0.0 & 13.3 & 0.294 & 0.188\\
         % & Debater                                      & 16.0 & 8.7 & 16.0 & 0.294 & 0.188\\
         & SMatchToPR (base)                            & 21.3 & 3.7 & 16.6 & 0.351 & 0.344\\
         & SMatchToPR (large)                           & 21.3 & 3.7 & 16.6 & 0.351 & 0.344\\
         \midrule
         \multirow{4}{*}{\datapersp} & ChatGPT           & 21.3 & 5.7 & 18.2 & 0.355 & 0.322\\
         & ChatGPT (closed book)                         & 17.1 & 3.8 & 15.0 & 0.291 & 0.258\\
         & Debater                                       & 9.4 & 0.4 & 8.5 & 0.197 & 0.210\\
        & SMatchToPR (base)                              & 22.5 & 6.5 & 19.3 & 0.257 & 0.232\\
        & SMatchToPR (large)                             & \textbf{22.7} & \textbf{6.7} & \textbf{19.4} & \textbf{0.403} & \textbf{0.363} \\
         \bottomrule
    \end{tabular}
    \caption{ROUGE scores and semantic similarity scores for the Key Point Generation task.}
    \label{tab:results-kpg}
\end{table*}

Table~\ref{tab:results-kpg} shows the results of KPG evaluation. 
Overall, no single approach performs best across all datasets. All models perform best on \datakp except for closed-book ChatGPT, which performs the best on the \datapve dataset. Thus, by adopting diverse datasets, we demonstrate that experimenting with a single dataset may inflate KPG performance. 

ChatGPT consistently scores well on ROUGE and semantic similarity. This indicates that the abstractive generation of key points is beneficial. For \datapve, we observe a strong tendency for open-book ChatGPT to adjust the generated key points to the linguistic style of the arguments. This clashes with the reference key points, which are paraphrased to make sense without the context of the original arguments. Hence, the closed-book model, which does not observe the source arguments, performs better, adopting more neutral language.

SMatchToPR performs best for \datapersp. Although general-purpose LLMs are strong in zero-shot settings, a dedicated model for representing arguments achieves state-of-the-art results. The Debater approach is ranked lowest across all datasets, showing that training on a single dataset generalizes poorly to other datasets.

ROUGE and semantic similarity scores mostly agree, except for BLEURT on \datakp. Here, we see that SMatchToPR slightly outperforms ChatGPT. 
We attribute this to the optimized representational qualities of SMatchToPR: it selects key points with high semantic similarity to many arguments, which is similar to how BLEURT provides scores based on contextualized representations.

Increasing model size (of SMatchToPR) improves performance for \datapersp, but not for \datakp and \datapve. 
Because \datapve is small, the pool of sentences to pick key point candidates from is limited, and possible improvements of the model are negligible when extracting the key points. For \datakp, the ROUGE scores deteriorate, while the semantic similarity scores improve slightly. Intuitively, this matches expectations: the model can navigate the embedding space better, selecting key points that may be phrased differently but contain semantically similar content.

\subsection{KPM Performance}
\label{sec:results-kpm}
Table~\ref{tab:results-kpm} shows the results of KPM evaluation. ChatGPT, despite its strong performance on KPG, does not accurately match arguments to key points. Interestingly, the Debater outperforms the SMatchToPR model on the \datakp dataset, but SMatchToPR is stronger on the \datapve and \datapersp datasets. SMatchToPR's strong performance on \datapersp and \datakp is expected--they were included in its training. However, its good performance on \datapve is interesting and it suggests that generalization is aided by more diverse data in training. 

\begin{table}[!htb]
    \centering
    \begin{tabular}{@{}l@{}ccc@{}}
         \toprule
         & \multicolumn{3}{c}{\textbf{mAP}}\\
         \cmidrule(lr){2-4}
         \textbf{Name}  & \datakp & \datapve & \textsc{Perspec.} \\
         \midrule
         ChatGPT &  0.17 & 0.27  & 0.46$^*$\\
         Debater & \textbf{0.82} & 0.51 & 0.51\\
         SMatchToPR (base) &  0.76 & 0.53 & 0.80 \\
         SMatchToPR (large) &  0.80 & \textbf{0.61} & \textbf{0.82} \\
         \bottomrule
    \end{tabular}
\caption{Results for the Key Point Matching task. Closed-book ChatGPT scores are not available, since its KPA is made without observing arguments. The scores for ChatGPT on \datapersp ($^*$) were estimated on a subset of the test set to cut down costs. }
    \label{tab:results-kpm}
\end{table}

\subsection{Analysis}
\paragraph{Long tail diversity}
Most key points and claims are heavily skewed in the number of data points, except for \datapve. Even for \datakp, where key points with few matching arguments were removed, there is a strong imbalance across claims and key points in terms of associated arguments (App.~\ref{fig:skewed}).

Following this imbalance, we sort key points by the number of associated arguments such that the least frequent key points are considered first. Then, we introduce a cutoff parameter $f$ to include arguments from a fraction of key points, starting with the least frequent. Using this parameter we perform matching only on low-frequency key point--arguments matches. This allows us to investigate the approaches' performance in the long tail.

\begin{figure}[htb]
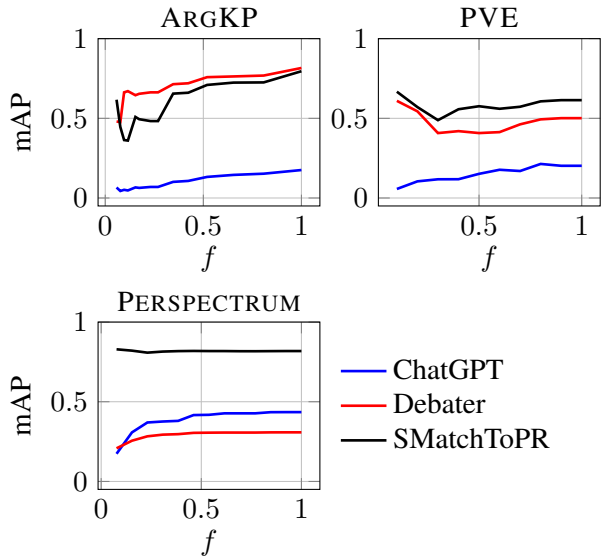

    \centering
    \includestandalone[mode=buildnew]{images/limit_kpm}
    \caption{KPM performance when limiting data usage to a fraction $f$, starting with long tail first.}
    \label{fig:results-limit}
\end{figure}

When we limit data usage by taking long tail arguments first, the performance of the KPA approaches, mainly on \datakp and \datapersp, decreases as shown in Figure~\ref{fig:results-limit}. 
This shows that the ability to correctly match arguments is contingent on the frequency of the arguments. In some cases, the arguments associated with key points with the fewest matches can be matched, but there is a strong performance loss for low values of $f$. Across all datasets, ChatGPT suffers consistently in mAP when conditioning on low-frequency key points. For SMatchToPR on \datapersp, there is almost no effect, showing that representation learning may positively impact the matching of key points to arguments even with low amounts of data.

Performing the same experiment for KPG results in similar results: key points with a low number of matched arguments are harder to represent well.

Next, we investigate whether the arguments in the long tail are different from the majority. Here, the long tail consists of arguments for key points that see less than the median number of arguments per key point.
We examine whether the sets of lexical items---noun phrase chunks (NPs) and entities---mentioned in the long tail arguments are included in the majority and vice versa. We also inspect the relative frequency of the shared lexical items via Kendall $\tau$ correlation on the NP and entity frequency rankings. Table~\ref{tab:lexical-analysis} shows these results. 

We see a large overlap of NPs and entities for \datakp between the long tail and the frequent key points. We attribute this to the filtering of low-frequency data during dataset construction. For the other two datasets, we observe much less overlap---in most cases, more than half of the noun phrases and entities are unique to either part of the dataset. The only exception here is \textsc{Perpectrum}, where roughly 40\% of the NPs and entities in the long tail are unique. When comparing the ranks of the intersecting lexical items, we observe moderate (but significant) rank correlation scores. Thus, the overlapping NPs and entities may not be in different frequencies in the two parts of the datasets. However, there is a strong indication of unique items in the long tail, in at least two of our datasets, showing that the long tail may contain novel insights. % next to the majority opinions. 

\begin{table*}[htb]
    \centering
    \begin{tabular}{@{}l@{\hspace{.3cm}}l@{\hspace{.2cm}}c@{\hspace{.2cm}}c@{\hspace{.2cm}}c@{\hspace{.2cm}}c@{\hspace{.3cm}}c@{\hspace{.2cm}}c@{}}
        \toprule
        & & \multicolumn{2}{c}{\textbf{NP}} & \multicolumn{2}{c}{\textbf{Entity}} & & \\
        \cmidrule(lr){3-4} \cmidrule(lr){5-6}
        \textbf{Left} & \textbf{Right} & left$-$right & right$-$left & left$-$right & right$-$left & NP-$\tau$ & Ent-$\tau$\\
        \midrule
        \datakp-long\_tail & \datakp-majority  & 0.168 & 0.234 & 0.191 & 0.273 & 0.216$^*$ & 0.373$^*$\\
        \datapve-long\_tail & \datapve-majority          & 0.638 & 0.787 & 0.719&  0.809 & 0.521$^*$ & 0.389 \\ 
        \textsc{Perspec.}-long\_tail & \textsc{Perspec.}-majority      & 0.397 & 0.807 & 0.401 & 0.797 & 0.361$^*$ & 0.427$^*$\\
        \bottomrule
    \end{tabular}
    \caption{Fraction of NPs and Entities in \textbf{Left} that are not in \textbf{Right} \& vice-versa. $^*$ indicates Kendall $\tau$ with $p<0.05$. }
    \label{tab:lexical-analysis}
\end{table*}

\paragraph{Annotator agreement}
Due to subjectivity in the annotation procedures, we expect annotators to rate argument--key point matches differently. We investigate whether the performance of KPA models reflects this subjectivity. That is, we test if match scores $x$ correlate with the agreement between annotators. Intuitively, when annotators agree, an argument and key point should be considered to match more objectively and thus may be easier to score for a model.  
From the two datasets that have a per-sample agreement score, we measure the Pearson $r$ correlation between the annotator agreement percentage (as obtained from data) and each approach's match score $M(d,k)$. Results are shown in Table~\ref{tab:results-annotators}.

\begin{table}[htb]
    \centering
    \begin{tabular}{@{}lcccc@{}}
        \toprule
         & \multicolumn{2}{c}{\datapve} & \multicolumn{2}{c}{\datapersp}\\  
         \cmidrule(lr){2-3} \cmidrule(lr){4-5}
        \textbf{Approach} & $r$ & $p$ & $r$ & $p$ \\
        \midrule
        ChatGPT & 0.030 & 0.687 & 0.039 & 0.469 \\
        Debater & 0.163 & 0.029& -0.051 & 0.013 \\
        SMatch-base & 0.097& 0.195 & 0.093 & 0.215 \\
        SMatch-large & 0.207 & 0.005 & -0.03 & 0.123 \\
        \bottomrule
    \end{tabular}
    \caption{Pearson $r$ correlation scores between predicted match scores and the annotator agreement per sample. }
    \label{tab:results-annotators}
\end{table}

For all approaches, the correlations are negligible or weak at best \citep{schober2018correlation}. 
This shows that the predictions made by the models fail to identify which matches are interpreted differently among annotators. Hence, these models are not able to represent the diversity stemming from annotation accurately \citep{plank2022problem}. 

\paragraph{Data sources} 
The KPG and KPM evaluations (Sections~\ref{sec:results-kpg} and~\ref{sec:results-kpm}) indicate how the methods perform when applied to different datasets. 
The performance is dataset- and task-specific; no single approach performs both tasks best on any dataset. 

\begin{figure}[htb]
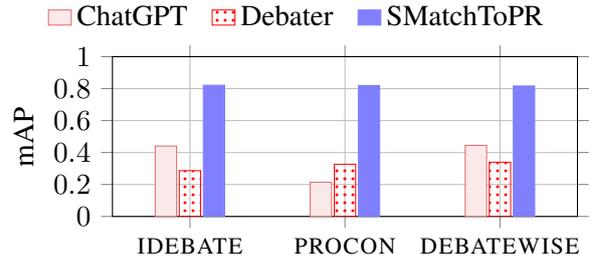

    \centering
    % Current scores show for SMatchToPR-large
    \includestandalone[mode=buildnew]{images/perspectrum_sources}
    \caption{KPM performance for all approaches on the different data sources in \datapersp.}
    \label{fig:perspectrum-sources}
\end{figure}

We further investigate the data sources in the \datapersp dataset, which was constructed using three distinct sources. Figure~\ref{fig:perspectrum-sources} shows the performance on each source separately. 
Although \datakp and \datapersp share a data source, we find no overlapping claims and little repetition in content between the two (App.~\ref{app:sec-data-details}).
 
The SMatchToPR and Debater approaches are not sensitive to data source shift, but ChatGPT performance differs depending on the source data used, dropping considerably for the \emph{procon} source. We find two factors that influence why these arguments are harder to match: 
\begin{enumerate*}[label=(\arabic*)]
    \item \emph{procon} contains about 10 times fewer claims than the other two sources, and
    \item \emph{procon}'s arguments are copied verbatim from various cited sources, leading to large stylistic and argumentative differences.
\end{enumerate*}

\section{Conclusion}
We perform a novel diversity exploration of different KPA approaches on three distinct datasets. By splitting KPA into two subtasks (KPG and KPM), we investigate each subtask, independently. 

First, we find that an LLM-based approach works well for generating key points, but fails to match arguments to key points reliably. Conversely, smaller fine-tuned models are better at matching arguments to key points but struggle to find good key points consistently. Second, using a single training set yields poor generalization across datasets, showing that data source impacts a KPA approach's ability to generalize. Diversification of training data leads to promising results. Third, across all datasets, we see that existing methods for KPA are insensitive to long tail diversity, decreasing performance for key points supported by few arguments. Finally, all models are insensitive to differences between individual annotators, disregarding subjective interpretations of arguments and key points.  

We showed how multiple aspects of diversity, a core principle when interpreting opinions, are not evaluated using the standard set of metrics.
Our analysis revealed interesting complementary strengths of the KPA approaches. Future efforts could focus on addressing diversity, either by mining for minority opinions directly \citep{waterschoot2022detecting}, or by identifying possibly subjective instances using socio-demographic information \citep{beck2023not}. 
Further, models can be enhanced with subjective understanding \citep{romberg2022your, vandermeer2023differences}, or work together with humans to jointly address some of the diversity issues \citep{van2022hyena, argyle2023leveraging}.

\section*{Ethics Statement}
There are growing ethical concerns about NLP (broadly, AI) technology, especially, when the technology is used in sensitive applications. Argument summarization can be used in sensitive applications, e.g., to assist in public policy making. An ethical scrutiny of such methods is necessary before their societal application. Our work contributes toward such scrutiny. The outcome of our analysis shows how KPA methods fail to handle diversity. Potential technological improvements may lead to better results, but due diligence is required before applying such methods to real-world use cases. 

We do not collect new data or involve human subjects in this work. Thus, we do not introduce any ethical considerations regarding data collection beyond those that affect the original datasets. A potential concern is that reproducing our results may involve using (possibly paid) services for running KPA. However, we aimed to make the analyses feasible with limited budget and resources.

\section*{Limitations}
We identify five limitations of our work.

\begin{description}[leftmargin=0pt]
\item [Diversity definition] Our definition of diversity is specific to three dimensions, but there may be additional dimensions. For example, our unit of analysis is at the \emph{argument} level. Diversity may also be analyzed for the opinion holders or those affected by decisions in policy-making contexts. 

\item[Novel key points] Our evaluation of KPG and KPM employs existing key points. However, KPA methods may generate novel or unseen key points. Evaluating such novel key points is nontrivial and it may require experiments involving human subjects.

\item[Resource limitations] KPA approaches are resource intensive. We limited some approaches where (1) it would become too expensive to run KPA because of the complexity of the number of comparisons (e.g., Debater approach), or (2) the models do not support a big enough window to fit all arguments (e.g., ChatGPT context window is limited). While there are alternatives (e.g., GPT-4), they drastically increase the cost. 

\item[Dataset diversity] The arguments in our data are in English, and limited to data gathered from online sources. Further, the users involved in collecting the datasets we employ may not be demographically representative of the global population. We conjecture that increasing the diversity of the data sources would make our conclusions stronger. However, publicly available datasets, especially non-English sources, for this task are scarce. We make our code and experimental data public to incentivize further research in this direction.

\item[Data exposure] We cannot verify whether the data from the test sets have been used when training the LLMs. This would make the model familiar with the vocabulary and have a more reliable estimation of the arguments' semantics. That likelihood is the smallest for \datapve since it is the most recent dataset, gathered with new crowd workers.
\end{description}

\section*{Acknowledgements}
This research was funded by the Netherlands Organisation for Scientific Research (NWO) through the Hybrid Intelligence Centre via the Zwaartekracht grant (024.004.022). We would like to thank the ARR reviewers for their feedback.

\bibliography{anthology,custom}
\bibliographystyle{acl_natbib}
\newpage
\appendix

\section{Detailed Experimental Setup}
We describe our experimental setup, starting with the data we use for conducting our analysis. We follow with a detailed description of each approach and finally present a description of the metrics used. 
\label{app:experimental-setup}
\subsection{Data}
\label{app:sec-data-details}
\begin{table*}[htb]
    \centering
    \begin{tabular}{@{}l
    >{\centering\arraybackslash}p{\dimexpr 0.15\linewidth-2\tabcolsep}
    >{\centering\arraybackslash}p{\dimexpr 0.15\linewidth-2\tabcolsep}
    >{\centering\arraybackslash}p{\dimexpr 0.15\linewidth-2\tabcolsep}
    >{\centering\arraybackslash}p{\dimexpr 0.15\linewidth-2\tabcolsep}
    >{\centering\arraybackslash}p{\dimexpr 0.15\linewidth-2\tabcolsep}@{}}
         \toprule
\textbf{Dataset} &  \makecell[c]{\textbf{Num.}\\\textbf{arguments}} & \makecell[c]{\textbf{Num.}\\\textbf{Key Points}} & \makecell[c]{\textbf{Num.}\\\textbf{claims}} & \makecell[c]{\textbf{Avg.}\\ \textbf{arguments }\\\textbf{per claim}} & \makecell[c]{\textbf{Avg.}\\\textbf{arguments }\\\textbf{per KP}}\\
\midrule
\datakp &  10717 & 277 & 31 & 245   & 20\\
\datapve &  269 & 185 & 3 & 67    & 4\\
\datapersp & 10927 & 3804 & 905   & 12    & 3\\
         \bottomrule
    \end{tabular}
    \caption{Quantitative statistics of the datasets used in the experiments.}
    \label{tab:datasets}
\end{table*}

\begin{figure}[htb]
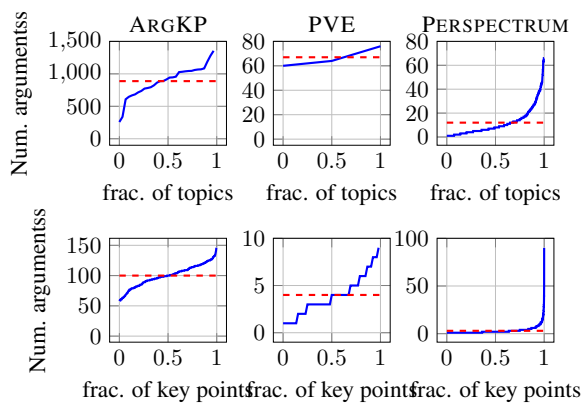

    \centering
    \includestandalone[mode=buildnew,width=\columnwidth]{images/long_tail}
    \caption{Number of arguments matched per claim (upper row) and key point (bottom row), sorted by frequency. The red dashed line shows the average number of arguments.}
    \label{fig:skewed}
\end{figure}

We provide some quantitative statistics on the three datasets used in our work in Table~\ref{tab:datasets}. \arrup{In addition, we show some qualitative examples of the content in our datasets in Table~\ref{app:tab:examples-qualitative}.} Since \datapersp and \datakp listed the same debate platforms as sources, we investigate the overlap between the claims and arguments between pairs of datasets. In terms of claims, there is no direct overlap between any two datasets. To rule out that the same arguments were scraped from the debate platforms, we also measure n-gram overlap \citep{clough2002measuring}. We show the overlap in unigrams, bigrams, and trigrams in Table~\ref{app:data-overlap}. The overlap scores report the ratio of n-grams from one dataset that is found in the other.

\begin{table*}[htb]
    \centering
    \begin{tabular}{@{}llccc@{}}
         \toprule
         & & \multicolumn{3}{c}{Target}\\
          \cmidrule(lr){3-5}
          && \datakp& \datapve & \datapersp  \\
         \midrule
         \multirow{3}{*}{Source} & \datakp & -- & 0.40/0.08/0.01 & 0.70/0.21/0.14 \\ 
         & \datapve & 0.41/0.16/0.06 & -- & 0.66/0.24/0.10 \\
         & \datapersp & 0.17/0.04/0.02& 0.22/0.03/0.01& -- \\
         \bottomrule
    \end{tabular}
    \caption{Maximum uni-, bi-, and trigram overlap between datasets.}
    \label{app:data-overlap}
\end{table*}

For \datapve, since the key point analysis was performed using a mixture of crowd and AI techniques, we take only the correctly matched key point--motivation pairs. That is, we take only those pairs that were deemed matching according to the final evaluation performed. 

\begin{table*}
    \centering
    \begin{tabular}{@{}lp{0.2\textwidth}p{0.3\textwidth}p{0.3\textwidth}@{}}
         \toprule
         \textbf{Dataset} & \textbf{Claim} & \textbf{Key Point} & \textbf{Argument} \\
         \midrule
         \datakp &  We should subsidize journalism & Journalism is important to information-spreading/accountability. & Journalism should be subsidized because democracy can only function if the electorate is well informed.\\
         \datapve & Young people may come together in small groups & Young people are at low risk of getting infected with COVID-19 and therefore can benefit from gathering together with limited risk and potential profit. & Risks of contamination or transfer have so far been found to be much smaller.\\
         \textsc{Persp.} & The threat of Climate Change is exaggerated & Overwhelming scientific consensus says human activity is primarily responsible for global climate change. & The biggest collection of specialist scientists in the world says that the world's climate is changing as a result of human activity. The scientific community almost unanimously agrees that man-caused global warming is a severe threat, and the evidence is stacking.\\
         \bottomrule
    \end{tabular}
    \caption{Qualtitative examples of claims, key points and arguments across our dataset.}
    \label{app:tab:examples-qualitative}
\end{table*}

\subsection{Per-approach Specifics}
See Table~\ref{app:tab:model-names} for the language models used in each approach. We further outline any details depending on the approach used. 
\begin{table}[htb]
    \centering
    \begin{tabular}{@{}lr@{}}
         \toprule
         \textbf{Name} & \textbf{Model} \\
         \midrule
         ChatGPT & \texttt{gpt-3.5-turbo-16k} \\
         \emph{ChatGPT (closed book)} & \texttt{gpt-3.5-turbo} \\
         Debater & \emph{closed-source} \\
         SMatchToPR (base)& \texttt{RoBERTa-base}\\
         SMatchToPR (large)& \texttt{RoBERTa-large}\\
         \bottomrule
    \end{tabular}
    \caption{Models used for each KPA approach.}
    \label{app:tab:model-names}
\end{table}

\paragraph{Debater} The Debater API allows multiple parameters when running the KPA analysis. We manually tuned the parameters separately for KPG and KPM. For both tasks, we started with the most permissive configuration to optimize for recall first, and gradually made parameters more strict to improve precision without lowering recall scores. Once recall scores started dropping, we fixed the parameters. The final configuration is shown in Table~\ref{app:tab:debater-params}.

\begin{table}[htb]
    \centering
    \begin{tabular}{@{}lr@{}}
         \toprule
         \textbf{Parameter} & \textbf{Value} \\
         \midrule
         \emph{KPG}& \\
         mapping\_policy & \emph{LOOSE} \\
         kp\_granularity & \emph{FINE}\\
         kp\_relative\_aq\_threshold & 0.5\\
         kp\_min\_len &  0 \\
         kp\_max\_len & 100\\
         kp\_min\_kp\_quality & 0.5\\
         \midrule
         \emph{KPM}\\
         min\_matches\_per\_kp & 0\\
         mapping\_policy & \emph{LOOSE}\\
         \bottomrule
    \end{tabular}
    \caption{API Configuration for Debater approach.}
    \label{app:tab:debater-params}
\end{table}

\paragraph{ChatGPT} 
We strive to make our results as reproducible as possible, but due to the nature of the OpenAI API results may be specific to model availability. We conducted the experiments between July and August 2023, using the \texttt{gpt-3.5-turbo} and \texttt{gpt-3.5-turbo-16k} models. We provide a template for the prompts below, in Prompts~\ref{prompt:gpt-closed},~\ref{prompt:gpt-open}, and~\ref{prompt:gpt-open-kpm}. Open-book ChatGPT for KPG uses up to $B_{KPG}={600, 100, 100}$ for \datakp, \datapve, \datapersp respectively. ChatGPT uses a batch size of $B_{KPM}={10}$ when making match predictions for KPM. 

Interpreting the responses was done by prompting the model to output valid JSON, and writing a script that parses the generated response. Invalid JSON responses are considered errors on the model's side, resulting in an empty string for KPG and a `no-match' label for KPM. In order to cut down on costs, we subsampled the test set for \datapersp, taking a random  $15\%$ of the claims in order to drive down the costs further.

\begin{theo}[ChatGPT closed book, KPG prompt]{prompt:gpt-closed}
Give me a JSON object of key arguments for and against the claim: \textbf{\{claim\}}. Make sure the reasons start with addressing the main point. Indicate per reason whether it supports (pro) or opposes (con) the claim. Rank all reasons from most to least popular. Make sure you generate a valid JSON object. The object should contain a list of dicts containing fields: 'reason' (str), 'popularity' (int), and 'stance' (str).
\end{theo}

\begin{theo}[ChatGPT open book, KPG prompt]{prompt:gpt-open}
Extract key arguments for and against the claim: \textbf{\{claim\}}. You need to extract the key arguments from the comments listed here: \textbf{\{up to $B_{KPG}$ arguments\}}
Give me a JSON object of key arguments for and against the claim. Make sure the reasons start with addressing the main point. Indicate per reason whether it supports (pro) or opposes (con) the claim. Rank all reasons from most to least popular. Make sure you generate a valid JSON object. The object should contain a list of dicts containing fields: 'reason' (str), 'popularity' (int), and 'stance' (str).
\end{theo}

\begin{theo}[ChatGPT open book, KPM prompt]{prompt:gpt-open-kpm}
For the claim of \textbf{\{claim\}}, indicate for each of the following argument/key point pairs whether the argument matches the key point. Return a JSON object with just a "match" boolean per argument/key point pair.

ID: \textbf{\{pair id\}} Argument: \textbf{\{argument\}} Key point: \textbf{\{key point\}}
 (\emph{up to $B_{KPM}$ times}) \dots
\end{theo}

\paragraph{SMatchToPR} We preprocess the \datapersp dataset analogously to the \datakp dataset. We train the SMatchToPR model using contrastive loss for 10 epochs and a batch size of 32. The training has a warmup phase of the first 10\% of data. The base and large variants use the same parameters. See Table~\ref{app:tab:smatch-params} for the hyperparameters when executing KPG and KPM. The computing infrastructure used contained two RTX3090 Ti GPUs. Training the RoBERTa large variant takes around 30 minutes. 

\begin{table}[htb]
    \centering
    \begin{tabular}{@{}lr@{}}
         \toprule
         \textbf{Parameter} & \textbf{Value} \\
         \midrule
            PR $d$ & 0.2 \\
            PR min quality score & 0.8 \\
            PR min match score & 0.8 \\
            PR min length & 5 \\
            PR max length & 20 \\
            filter min match score & 0.5 \\ 
            filter min result length & 5\\
            filter timeout & 1000\\
         \bottomrule
    \end{tabular}
    \caption{Hyperparameters for SMatchToPR approach.}
    \label{app:tab:smatch-params}
\end{table}

% Number of parameters
% Total computational budget
% Computing infra

\subsection{Evaluation metrics}
\label{app:sec-rouge}
\paragraph{KPG} Since we use ROUGE scores for evaluation, to make our results reproducible we provide further details on the configuration of the ROUGE metrics \citep{grusky2023rogue}. Our evaluation uses the \texttt{sacrerouge} package that wraps the original ROUGE implementation\footnote{\url{https://github.com/danieldeutsch/sacrerouge}}. The full evaluation parameters can be seen in Table~\ref{app:tab:rouge-params}.
\begin{table}[htb]
    \centering
    \begin{tabular}{@{}lr@{}}
         \toprule
         \textbf{Parameter} & \textbf{Value} \\
         \midrule
         Porter Stemmer & \emph{yes} \\
         Confidence Interval & 95\\
         Bootstrap samples & 1000\\
         $\alpha$ & 0.5\\
         Counting unit & \emph{sentence}\\
         \bottomrule
    \end{tabular}
    \caption{Configuration parameters for the ROUGE evaluation of KPG.}
    \label{app:tab:rouge-params}
\end{table}

Furthermore, we use two learned metrics (BLEURT and BARTScore) to report the semantic similarity of generated key points and reference key points. For BLEURT, we use the publicly available \texttt{BLEURT-20} model, which is a RemBERT \citep{chung2020rethinking} model trained on an augmented version of the WMT shared task data \citep{ma2019results}. BARTScore uses a BART model trained on ParaBank2 \citep{hu2019large}.

\section{Additional results}
We present two additional results: we provide fine-grained ROUGE results for KPG, and provide examples of key points generated by ChatGPT. 
\label{sec:appendix-additional}
\subsection{Detailed ROUGE scores for Key Point Generation}
Earlier, we provided aggregated $F_1$ scores for the KPG evaluation. Here, we also show Precision and Recall scores in Table~\ref{app:tab:results-kpg-fine}. We see that the models that perform best in terms of $F_1$ score are consistently scoring well in terms of precision and recall across all datasets. For instance, open-book ChatGPT performs best on \datakp in terms of $F_1$ (see Table~\ref{tab:results-kpg}), achieving consistently high precision and recall scores. Other approaches may score higher on individual metrics (e.g. SMatchToPR large scores higher in terms of ROUGE-1 recall), but this pattern is not consistent across all metric types. 

\begin{table*}[!htb]
    \centering
    \begin{tabular}{@{}llcccccc@{}}
         \toprule
         & &\multicolumn{3}{c}{Precision} &  \multicolumn{3}{c}{Recall}\\
         \cmidrule(lr){3-5} \cmidrule(lr){6-8}
        \textbf{Data} & \textbf{Approach} & \textbf{R-1} & \textbf{R-2} & \textbf{R-L} & \textbf{R-1} & \textbf{R-2} & \textbf{R-L} \\
         \midrule
         \multirow{4}{*}{\datakp} & ChatGPT             & 29.1 & \textbf{10.6} & 25.6  & 45.2 & \textbf{16.1} & 41.2 \\
         & ChatGPT (closed book)                        & \textbf{30.8} & 6.8 & \textbf{26.9}  & 32.0 & 8.6 & 27.3 \\
         & Debater                                      & 25.3 & 5.5 & 23.1  & 28.2 & 5.3 & 23.4 \\
         & SMatchToPR (base)                            & 24.5 & 9.3 & 23.2  & 44.5 & 11.2 & 41.5 \\
         & SMatchToPR (large)                           & 22.0 & 6.4 & 19.4 & \textbf{53.0} & 13.0 & \textbf{47.5}\\
         \midrule
         \multirow{4}{*}{\datapve} &          ChatGPT   & 25.1 & 6.4 & 21.1 & 19.1 & 3.9 & 15.8\\
         & ChatGPT (closed book)                        & 30.1 & \textbf{9.8} & 22.6 & \textbf{26.4} & \textbf{8.1} & \textbf{21.6}\\
         & Debater                                      & \textbf{33.3} & 0.0 & \textbf{33.3}  & 13.3 & 7.1 & 13.3 \\
         & SMatchToPR (base)                            & 28.8 & 5.6 & 22.6  & 18.0 & 2.9 & 14.4 \\
         & SMatchToPR (large)                           & 27.8 & 5.6 & 22.6 & 18.0 & 2.9 & 14.4\\
         \midrule
         \multirow{4}{*}{\datapersp} & ChatGPT          & 17.5 & 4.7 & 14.8 & \textbf{35.0} & \textbf{10.2} & \textbf{30.5}\\
         & ChatGPT (closed book)                        & 14.8 & 3.1 & 12.8  & 25.4 & 6.3 & 22.7 \\
         & Debater                                      & 8.6 & 0.4 & 7.6  & 25.5 & 6.3 & 22.7 \\
        & SMatchToPR (base)                             & 18.8 & 5.5 & 15.9  & 32.0 & 9.2 & 27.8 \\
        & SMatchToPR (large)                            & \textbf{19.0} & \textbf{5.7} & \textbf{16.1}  & 32.3 & 9.8 & 28.3 \\
         \bottomrule
    \end{tabular}
    \caption{ROUGE Precision and Recall scores for the Key Point Generation task.}
    \label{app:tab:results-kpg-fine}
\end{table*}

\subsection{Additional BERTScores for Key Point Generation}
\arrup{Next to BLEURT and BARTScore, we report BERTScore \citep{bertscore2020} for the approaches in the KPG evaluation, to examine the relation between the various learned metrics. }
\begin{table*}[!htb]
    \centering
    \begin{tabular}{@{}llccc@{}}
         \toprule
         & &\multicolumn{3}{c}{BERTScore} \\
         \cmidrule(lr){3-5}
        \textbf{Data} & \textbf{Approach} & \textbf{Precision} & \textbf{Recall} & $\mathbf{F}_\mathbf{1}$\\
         \midrule
         \multirow{4}{*}{\datakp} & ChatGPT             & 0.412 & 0.470& 0.422\\
         & ChatGPT (closed book)                        & 0.322 & 0.336& 0.324\\
         & Debater                                      & 0.406 & 0.367 & 0.379\\
         & SMatchToPR (base)                            & 0.362 & 0.463 & 0.394\\
         & SMatchToPR (large)                           & 0.361& 0.482& 0.402\\
         \midrule
         \multirow{4}{*}{\datapve} &          ChatGPT   & 0.184 & 0.157 & 0.153\\
         & ChatGPT (closed book)                        & 0.386 & 0.280  & 0.324\\
         & Debater                                      & 0.523 & 0.146 & 0.301\\
         & SMatchToPR (base)                            & 0.339& 0.210& 0.257\\
         & SMatchToPR (large)                           & 0.339& 0.210& 0.257\\
         \midrule
         \multirow{4}{*}{\datapersp} & ChatGPT          & 0.208 & 0.308 & 0.252\\
         & ChatGPT (closed book)                        & 0.244 & 0.274 & 0.243\\
         & Debater                                      & 0.228 & 0.274 &0.246\\
        & SMatchToPR (base)                             & 0.231 & 0.297 & 0.258\\
        & SMatchToPR (large)                            & 0.235 & 0.296 & 0.260\\
         \bottomrule
    \end{tabular}
    \caption{BERTScore Precision, Recall, and $F_1$ scores for the Key Point Generation task.}
    \label{app:tab:results-kpg-bertscore}
\end{table*}

\subsection{Long-tail experiment for KPG}
\arrup{We perform the long-tail analysis for Key Point Generation, adopting the same cutoff parameter $f$ from the KPM analysis. Figure~\ref{app:fig:long-tail-kpg} shows the results when including a fraction of key points $f$, starting from the least frequent (i.e. the key points with the lowest amount of arguments matched to them). The figure shows that for a low fraction of data, all approaches perform considerably worse. Note that due to the evaluation setup in \citet{li2023hear}, scores may be lower due to a smaller pool of key points. Since we report averages of the maximum scoring match between any given generated and reference key points, this smaller pool may lead to overall lower scores. We still report these results to show the impact of making the evaluation set smaller, next to focusing on infrequent opinions.}

\begin{figure*}
    \centering
    \includestandalone[width=\textwidth]{images/limit_kpg}
    \caption{KPG performance when limiting data usage to a fraction $f$, starting with the long tail first.}
    \label{app:fig:long-tail-kpg}
\end{figure*}

\subsection{ChatGPT generated key points for \datapve} 
See Table~\ref{app:tab:examples-chatgpt}. A cursory search for the content of the open-book key points shows the key points are directly taken from arguments in \datapve. While ChatGPT performs conditioned language generation, it behaves like extractive summarization when using the open-book approach for the arguments in \datapve. This leads to potentially incomplete or subjective key points. For the closed-book approach, we observe that ChatGPT generates independent and objective key points.
\begin{table*}[htb]
    \centering
    \begin{tabular}{@{}p{3cm}lp{4cm}p{5cm}@{}}
         \toprule
        \textbf{Claim} & \textbf{Stance} & \textbf{KP (open-book)} & \textbf{KP (closed-book)}\\
         \midrule
          All restrictions are lifted for persons who are immune & con & The coronavirus is an assassin, let's really learn more about this first & There may still be unknown long-term effects of the virus, even in those who have recovered.\\
         Re-open hospitality and entertainment industry  & pro & Economy needs to start running again  & Reopening the hospitality and entertainment industry will help stimulate the economy and create job opportunities.\\
         Young people may come together in small groups & con & The spread will then come back in all its intensity. & Small group gatherings may pose a risk of spreading contagious diseases.\\
         \bottomrule
    \end{tabular}
    \caption{Examples of generated key points from the open-book and closed-book ChatGPT approach. }
    \label{app:tab:examples-chatgpt}
\end{table*}

% \subsection{Key Point Generation with expert-generated key points}
% In addition to key points contained in the \datapve corpus, we can use the human-generated key points from a manual analysis of 600 comments \citep{mouter2021public}. The results can be seen in Table~\ref{tab:results-kpg-hyena-manual}.
% \begin{table*}[htb]
%     \centering
%     \begin{tabular}{@{}lccccc@{}}
%          \toprule
%          & & \multicolumn{4}{c}{\datapve (Manual)} \\
%          \textbf{Name} & \textbf{R-1} & \textbf{R-2} & \textbf{R-L} & \textbf{BLEURT (sF1) } & \textbf{BART (sF1)} \\
%          \midrule
%          ChatGPT &  22.1 & 3.9 & 18.3 & 0.362 & \textbf{0.382}\\
%          ChatGPT (closed book)&  \textbf{25.6} & \textbf{8.5} & \textbf{22.9} & \textbf{0.395} & 0.365\\
%          Debater & 6.7 & 0.0 & 6.7 & 0.212 & 0.176\\
%          SMatchToPR-base & 23.4 & 3.0 & 19.0 & 0.337 & 0.346\\
%          SMatchToPR-large & 23.4 & 3.0 & 19.0 & 0.337 & 0.346\\
%          \bottomrule
%     \end{tabular}
%     \caption{Results for the Key Point Generation task for manually generated key points in \datapve from \citet{mouter2021public}.}
%   \label{tab:results-kpg-hyena-manual}
% \end{table*}

\end{document}

%% file: macros.tex
\usepackage{xcolor}
\usepackage{amssymb}

\definecolor{leacolor}{rgb}{0.40,0.60,1.00}
\definecolor{selcolor}{rgb}{0.53,0.30,1.00}
\definecolor{mmcolor}{rgb}{0.1,0.2,0.4}
\definecolor{urjacolor}{rgb}{0.039, 0.478, 0.741}

\definecolor{todocolor}{rgb}{1.00,0.75,0.00}
\definecolor{changedcolor}{rgb}{0.42,0.27,0.57}
\definecolor{removedcolor}{rgb}{0.867,0.176,0.361}
\definecolor{removedcolor}{rgb}{0.867,0.176,0.361}

\newcommand{\arrupdate}[2]{
    {\textcolor{#2}{#1}}
}

\newcommand{\arrup}[1]{\arrupdate{#1}{black}}

%% file: images/limit_kpm.tex
    

\begin{tikzpicture}
    \begin{axis}[
        name=topic-kpa2021,
        title = {\textsc{ArgKP}},
        clip=false,
        anchor=west,
        height=3.8cm,
        width=4.5cm,
        title style={yshift=-0.2cm},
        xlabel = {$f$},
        ylabel = {mAP},
        xlabel style={yshift=0.3em},
        ymax=1.0,
        ymin=-0.05,
        grid = major,
        grid style={line width=.1pt, draw=gray!50},
    ]
    \addplot[
        blue,
        line width=1pt,
    ] table[x=x, y=chatgpt]{images/data/limit_perf_kpa2021.dat}; 
    \addplot[
        red,
        line width=1pt,
    ] table[x=x, y=kpa_api]{images/data/limit_perf_kpa2021.dat}; 
    \addplot[
        black,
        line width=1pt,
    ] table[x=x, y=smatch]{images/data/limit_perf_kpa2021.dat}; 
    \end{axis}
    
    \begin{axis}[
        name=topic-hyena,
        at={($(topic-kpa2021.east)+(2em, 0)$)},
        title = {\textsc{PVE}},
        clip=false,
        height=3.8cm,
        width=4.5cm,
        anchor=west,
        title style={yshift=-0.2cm},
        xlabel = {$f$},
        xlabel style={yshift=0.3em},
        grid = major,
        ymax=1.0,
        ymin=-0.05,
        grid style={line width=.1pt, draw=gray!50},
    ]
    \addplot[
        blue,
        line width=1pt,
    ] table[x=x, y=chatgpt]{images/data/limit_perf_hyena.dat}; 
    \addplot[
        red,
        line width=1pt,
    ] table[x=x, y=kpa_api]{images/data/limit_perf_hyena.dat}; 
    \addplot[
        black,
        line width=1pt,
    ] table[x=x, y=smatch]{images/data/limit_perf_hyena.dat}; 
    \end{axis}
    
    \begin{axis}[
        name=topic-perspectrum,
        at={($(topic-kpa2021.south)+(0em, -4em)$)},
        clip=false,
        height=3.8cm,
        width=4.5cm,
        anchor=north,
        title = {\textsc{Perspectrum}},
        ylabel={mAP},
        title style={yshift=-0.2cm},
        xlabel = {$f$},
        xlabel style={yshift=0.3em},
        grid = major,
        ymax=1.0,
        ymin=-0.05,
        grid style={line width=.1pt, draw=gray!50},
        legend cell align=left,
        legend style={
            at={(1.05,0.5)},
            draw=none,
            anchor=west,
            /tikz/every even column/.append style={
                column sep=0.25cm
            }
        },
        legend columns=1,
    ]
    \addplot[
        blue,
        line width=1pt,
    ] table[x=x, y=chatgpt]{images/data/limit_perf_perspectrum.dat}; 
    \addplot[
        red,
        line width=1pt,
    ] table[x=x, y=kpa_api]{images/data/limit_perf_perspectrum.dat}; 
    \addplot[
        black,
        line width=1pt,
    ] table[x=x, y=smatch]{images/data/limit_perf_perspectrum.dat}; 
    \legend{ChatGPT, Debater, SMatchToPR};
    \end{axis}

\end{tikzpicture}

%% file: images/perspectrum_sources.tex
\begin{tikzpicture}
    \begin{axis} [ybar,
        bar width = 8pt,
        width=\columnwidth,
        height=3.7cm,
        xtick = data,
        ymin = 0,
        ymax = 1,
        grid = major,
        ytick align=outside,
        xticklabels={\textsc{idebate}, \textsc{procon}, \textsc{debatewise}},
        ylabel={mAP},
        legend cell align=left,
        legend style={
            at={(0.4,1.1)},
            legend columns=3,
            draw=none,
            anchor=south,
            /tikz/every even column/.append style={
                column sep=0.25cm
            }
        },
        legend image code/.code={%
          \draw[#1] (0cm,-0.1cm) rectangle (0.3cm,0.1cm);
        },
        enlarge x limits = {.25},
    ]
        \addplot+[draw = red!70!gray!70,
            line width = 0.1mm,
            fill = red!70!gray!10,
            bar shift=-9pt,
        ] coordinates {(1, 0.4406) (2, 0.2143) (3,  0.4452)};

        \addplot [draw = red,
            line width = 0.1mm,
            pattern = dots,
            pattern color = red,
            bar shift=0pt,
        ] coordinates {(1, 0.2865) (2, 0.3269) (3, 0.3396)};

        \addplot [draw = blue!50,
            line width = 0.1mm,
            fill = blue!50,
            bar shift=9pt,
        ]   coordinates {(1,0.8214) (2,   0.8192) (3, 0.8166)};
        \legend {ChatGPT, Debater, SMatchToPR};
    \end{axis}

\end{tikzpicture}

%% file: images/long_tail.tex
\begin{tikzpicture}
    \begin{axis}[
        name=topic-kpa2021,
        title = {\textsc{ArgKP}},
        clip=false,
        anchor=west,
        width=3.5cm,
        height=3.3cm,
        title style={yshift=-0.2cm},
        xlabel = {frac. of topics},
        ylabel = {Num. argumentss},
        ymin=-100,
        grid = major,
        grid style={line width=.1pt, draw=gray!50},
    ]
    \addplot[
        blue,
        line width=1pt,
    ] table {images/data/long_tail_topics_kpa2021.dat};
    \addplot[
        red,
        line width=1pt,
        dashed,
    ] coordinates {
        (0,888)
        (1,888)
    }; 
    \end{axis}
    
    \begin{axis}[
        name=topic-hyena,
        at={($(topic-kpa2021.east)+(2em, 0)$)},
        title = {\textsc{PVE}},
        clip=false,
        height=3.3cm,
        width=3.5cm,
        anchor=west,
        title style={yshift=-0.2cm},
        % width=0.5\columnwidth,
        % height=5cm,
        xlabel = {frac. of topics},
        ymin=-5,
        ymax=80,
        grid = major,
        grid style={line width=.1pt, draw=gray!50},
    ]
    \addplot[
        blue,
        line width=1pt,
    ] table {images/data/long_tail_topics_hyena.dat};
    \addplot[
        red,
        line width=1pt,
        dashed,
    ] coordinates {
        (0,67)
        (1,67)
    }; 
    \end{axis}
    
    \begin{axis}[
        name=topic-perspectrum,
        at={($(topic-hyena.east)+(2em, 0)$)},
        title = {\textsc{Perspectrum}},
        clip=false,
        height=3.3cm,
        width=3.5cm,
        anchor=west,
        title style={yshift=-0.2cm},
        % width=0.5\columnwidth,
        % height=5cm,
        xlabel = {frac. of topics},
        ymax=80,
        grid = major,
        grid style={line width=.1pt, draw=gray!50},
    ]
    \addplot[
        blue,
        line width=1pt,
    ] table {images/data/long_tail_topics_perspectrum.dat};
    \addplot[
        red,
        line width=1pt,
        dashed,
    ] coordinates {
        (0,12)
        (1,12)
    }; 
    \end{axis}

    \begin{axis}[
        name=kp-kpa2021,
        clip=false,
        at={($(topic-kpa2021.south)+(0em, -4em)$)},
        anchor=north,
        height=3.3cm,
        width=3.5cm,
        title style={yshift=-0.2cm},
        xlabel = {frac. of key points},
        ylabel = {Num. argumentss},
        ymin=-10,
        grid = major,
        grid style={line width=.1pt, draw=gray!50},
    ]
    \addplot[
        blue,
        line width=1pt,
    ] table {images/data/long_tail_kp_kpa2021.dat};
    \addplot[
        red,
        line width=1pt,
        dashed,
    ] coordinates {
        (0,100)
        (1,100)
    }; 
    \end{axis}

    \begin{axis}[
        name=kp-hyena,
        clip=false,
        at={($(kp-kpa2021.east)+(2em, 0em)$)},
        anchor=west,
        height=3.3cm,
        width=3.5cm,
        title style={yshift=-0.2cm},
        xlabel = {frac. of key points},
        ymin=-1,
        grid = major,
        grid style={line width=.1pt, draw=gray!50},
    ]
    \addplot[
        blue,
        line width=1pt,
    ] table {images/data/long_tail_kp_hyena.dat};
    \addplot[
        red,
        line width=1pt,
        dashed,
    ] coordinates {
        (0,4)
        (1,4)
    }; 
    \end{axis}

    \begin{axis}[
        name=kp-perspectrum,
        clip=false,
        at={($(kp-hyena.east)+(2em, 0em)$)},
        anchor=west,
        height=3.3cm,
        width=3.5cm,
        title style={yshift=-0.2cm},
        xlabel = {frac. of key points},
        ymax=100,
        grid = major,
        xtick={0, 0.5,1.0},
        grid style={line width=.1pt, draw=gray!50},
    ]
    \addplot[
        blue,
        line width=1pt,
    ] table {images/data/long_tail_kp_perspectrum.dat};
    \addplot[
        red,
        line width=1pt,
        dashed,
    ] coordinates {
        (0,3)
        (1,3)
    }; 
    \end{axis}
\end{tikzpicture}

%% file: images/limit_kpg.tex
    

\begin{tikzpicture}
    \begin{axis}[
        name=topic-kpa2021,
        title = {\textsc{ArgKP}},
        clip=false,
        anchor=west,
        height=5cm,
        width=5.5cm,
        ymin=0,
        ymax=14,
        title style={yshift=-0.2cm},
        xlabel = {$f$},
        ylabel = {ROUGE-2},
        xlabel style={yshift=0.3em},
        grid = major,
        grid style={line width=.1pt, draw=gray!50},
    ]
    \addplot[
        blue,
        line width=1pt,
    ] table[x=x, y=chatgpt_open]{images/data/limit_kpg_kpa2021_rouge2.dat}; 
    \addplot[
        blue,
        dashed,
        line width=1pt,
    ] table[x=x, y=chatgpt_closed]{images/data/limit_kpg_kpa2021_rouge2.dat}; 
    \addplot[
        red,
        line width=1pt,
    ] table[x=x, y=kpa_api]{images/data/limit_kpg_kpa2021_rouge2.dat}; 
    \addplot[
        black,
        line width=1pt,
    ] table[x=x, y=smatch]{images/data/limit_kpg_kpa2021_rouge2.dat}; 
    \end{axis}
    
    \begin{axis}[
        name=topic-hyena,
        at={($(topic-kpa2021.east)+(2em, 0)$)},
        title = {\textsc{PVE}},
        clip=false,
        height=5cm,
        width=5.5cm,
        ymin=0,
        ymax=14,
        anchor=west,
        title style={yshift=-0.2cm},
        xlabel = {$f$},
        xlabel style={yshift=0.3em},
        grid = major,
        grid style={line width=.1pt, draw=gray!50},
    ]
    \addplot[
        blue,
        line width=1pt,
    ] table[x=x, y=chatgpt_open]{images/data/limit_kpg_hyena_rouge2.dat}; 
    \addplot[
        blue,
        dashed,
        line width=1pt,
    ] table[x=x, y=chatgpt_closed]{images/data/limit_kpg_hyena_rouge2.dat}; 
    \addplot[
        red,
        line width=1pt,
    ] table[x=x, y=kpa_api]{images/data/limit_kpg_hyena_rouge2.dat}; 
    \addplot[
        black,
        line width=1pt,
    ] table[x=x, y=smatch]{images/data/limit_kpg_hyena_rouge2.dat}; 
    \end{axis}
    
    \begin{axis}[
        name=topic-perspectrum,
        at={($(topic-hyena.east)+(2em, 0em)$)},
        clip=false,
        height=5cm,
        width=5.5cm,
        ymin=0,
        ymax=14,
        anchor=west,
        title = {\textsc{Perspectrum}},
        title style={yshift=-0.2cm},
        xlabel = {$f$},
        xlabel style={yshift=0.3em},
        grid = major,
        grid style={line width=.1pt, draw=gray!50},
        legend cell align=left,
        legend style={
            at={(-0.8,1.2)},
            draw=none,
            anchor=south,
            /tikz/every even column/.append style={
                column sep=0.25cm
            }
        },
        legend columns=4,
    ]
    \addplot[
        blue,
        line width=1pt,
    ] table[x=x, y=chatgpt_open]{images/data/limit_kpg_perspectrum_rouge2.dat}; 
    \addplot[
        blue,
        dashed,
        line width=1pt,
    ] table[x=x, y=chatgpt_closed]{images/data/limit_kpg_perspectrum_rouge2.dat}; 
    \addplot[
        red,
        line width=1pt,
    ] table[x=x, y=kpa_api]{images/data/limit_kpg_perspectrum_rouge2.dat}; 
    \addplot[
        black,
        line width=1pt,
    ] table[x=x, y=smatch]{images/data/limit_kpg_perspectrum_rouge2.dat}; 
    \legend{ChatGPT (open), ChatGPT (closed), Debater, SMatchToPR};
    \end{axis}

% ======= ROW 2 ===================

    \begin{axis}[
        name=bart-kpa2021,
        at={($(topic-kpa2021.south)+(0em, -3em)$)},
        clip=false,
        anchor=north,
        height=5cm,
        width=5.5cm,
        ymin=0.15,
        ymax=0.55,
        title style={yshift=-0.2cm},
        xlabel = {$f$},
        ylabel = {BARTScore},
        xlabel style={yshift=0.3em},
        grid = major,
        grid style={line width=.1pt, draw=gray!50},
    ]
    \addplot[
        blue,
        line width=1pt,
    ] table[x=x, y=chatgpt_open]{images/data/limit_kpg_kpa2021_bart_f1.dat}; 
    \addplot[
        blue,
        dashed,
        line width=1pt,
    ] table[x=x, y=chatgpt_closed]{images/data/limit_kpg_kpa2021_bart_f1.dat}; 
    \addplot[
        red,
        line width=1pt,
    ] table[x=x, y=kpa_api]{images/data/limit_kpg_kpa2021_bart_f1.dat}; 
    \addplot[
        black,
        line width=1pt,
    ] table[x=x, y=smatch]{images/data/limit_kpg_kpa2021_bart_f1.dat}; 
    \end{axis}
    
    \begin{axis}[
        name=bart-hyena,
        at={($(bart-kpa2021.east)+(2em, 0)$)},
        clip=false,
        height=5cm,
        width=5.5cm,
        ymin=0.15,
        ymax=0.55,
        anchor=west,
        title style={yshift=-0.2cm},
        xlabel = {$f$},
        xlabel style={yshift=0.3em},
        grid = major,
        grid style={line width=.1pt, draw=gray!50},
    ]
    \addplot[
        blue,
        line width=1pt,
    ] table[x=x, y=chatgpt_open]{images/data/limit_kpg_hyena_bart_f1.dat}; 
    \addplot[
        blue,
        dashed,
        line width=1pt,
    ] table[x=x, y=chatgpt_closed]{images/data/limit_kpg_hyena_bart_f1.dat}; 
    \addplot[
        red,
        line width=1pt,
    ] table[x=x, y=kpa_api]{images/data/limit_kpg_hyena_bart_f1.dat}; 
    \addplot[
        black,
        line width=1pt,
    ] table[x=x, y=smatch]{images/data/limit_kpg_hyena_bart_f1.dat}; 
    \end{axis}
    
    \begin{axis}[
        name=bart-perspectrum,
        at={($(bart-hyena.east)+(2em, 0em)$)},
        clip=false,
        height=5cm,
        width=5.5cm,
        ymin=0.15,
        ymax=0.55,
        anchor=west,
        title style={yshift=-0.2cm},
        xlabel = {$f$},
        xlabel style={yshift=0.3em},
        grid = major,
        grid style={line width=.1pt, draw=gray!50},
        legend cell align=left,
        legend style={
            at={(-0.8,1.2)},
            draw=none,
            anchor=south,
            /tikz/every even column/.append style={
                column sep=0.25cm
            }
        },
        legend columns=4,
    ]
    \addplot[
        blue,
        line width=1pt,
    ] table[x=x, y=chatgpt_open]{images/data/limit_kpg_perspectrum_bart_f1.dat}; 
    \addplot[
        blue,
        dashed,
        line width=1pt,
    ] table[x=x, y=chatgpt_closed]{images/data/limit_kpg_perspectrum_bart_f1.dat}; 
    \addplot[
        red,
        line width=1pt,
    ] table[x=x, y=kpa_api]{images/data/limit_kpg_perspectrum_bart_f1.dat}; 
    \addplot[
        black,
        line width=1pt,
    ] table[x=x, y=smatch]{images/data/limit_kpg_perspectrum_bart_f1.dat}; 
    \end{axis}

\end{tikzpicture}